\documentclass{article} 
\usepackage{nips13submit_e,times}
\usepackage{times}

\usepackage{epsfig}
\usepackage{graphicx}
\usepackage{epstopdf}
\usepackage{amsmath}
\usepackage{amssymb}
\usepackage{mathtools}
\usepackage[numbers]{natbib}
\usepackage{subcaption}
\usepackage[font=small]{caption}

\usepackage[linktocpage=true,plainpages=false,pdfpagelabels=false]{hyperref}
\definecolor{darkblue}{rgb}{0,0.1,0.5}
\hypersetup{
    colorlinks, linkcolor={darkblue},
    citecolor={darkblue}, urlcolor={darkblue}
}
\usepackage{url}

\title{Multi-utility Learning: \\ Structured-output Learning with \\ Multiple Annotation-specific Loss Functions }

\author{
Roman Shapovalov \quad\quad Dmitry Vetrov \quad\quad Anton Osokin \\
Lomonosov Moscow State University\\
\texttt{shapovalov@graphics.cs.msu.ru} \\
\texttt{vetrovd@yandex.ru} \quad \texttt{anton.osokin@gmail.com} \\
\And
Pushmeet Kohli \\
Microsoft Research Cambridge \\
\texttt{pkohli@microsoft.com} \\
}

\def\vec#1{\mathbf#1}
\def\Kappa{\mathrm{K}}

\DeclareMathOperator*{\argmax}{\arg\!\max}

\nipsfinalcopy 

\begin{document}

\maketitle

\begin{abstract}
Structured-output learning is a challenging problem; particularly so because of the difficulty in obtaining large datasets of fully labelled instances for training. In this paper we try to overcome this difficulty by presenting a multi-utility learning framework for structured prediction that can learn from training instances with different forms of supervision. We propose a unified technique for inferring the loss functions most suitable for quantifying the consistency of solutions with the given weak annotation. We demonstrate the effectiveness of our framework on the challenging semantic image segmentation problem for which a wide variety of annotations can be used.  For instance, the popular training datasets for semantic segmentation are composed of images with hard-to-generate full pixel labellings, as well as images with easy-to-obtain weak annotations, such as bounding boxes around objects, or image-level labels that specify which object categories are present in an image. Experimental evaluation shows that the use of annotation-specific loss functions dramatically improves segmentation accuracy compared to the baseline system where only one type of weak annotation is used. 
\end{abstract}

\section{Introduction}
\begin{figure}[b]
\begin{center}
    \begin{minipage}[b]{.24\linewidth}
        \includegraphics[scale=0.4]{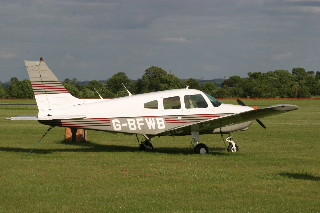}
        \subcaption{Original image}
        \label{fig:MSRC-orig}
    \end{minipage}
    \begin{minipage}[b]{.24\linewidth}
        \includegraphics[scale=0.4]{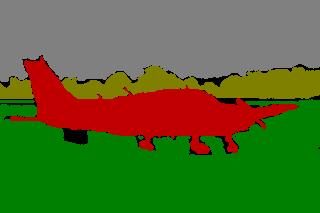}
        \subcaption{Full (strong) labelling}
        \label{fig:MSRC-strong}
    \end{minipage}
    \begin{minipage}[b]{.24\linewidth}
        \includegraphics[scale=0.4]{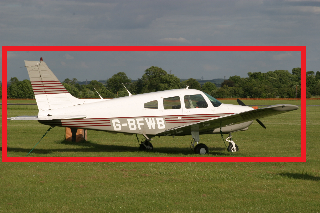}
        \subcaption{Bnd.-box annotation}
        \label{fig:MSRC-bbox}
    \end{minipage}
    \begin{minipage}[b]{.24\linewidth}
        \includegraphics[scale=0.4]{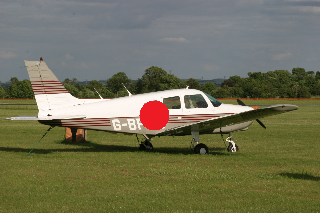}
        \subcaption{Object-seed annotation}
        \label{fig:MSRC-seed}
    \end{minipage}
    \caption{ Types of annotation for an image from the MSRC dataset~\cite{Shotton2006}}
    \label{fig:MSRC-label-types}
\end{center}
\end{figure}
Training structured-output classifiers is a challenging problem; not only because of the associated computational burden, but also due to difficulties in obtaining the ground-truth labelling for training data: in problems like semantic image segmentation the structured label may comprise thousands of scalars, so annotation of large datasets requires a lot of human effort. In contrast, it is much easier to obtain a weak annotation of an image, i.e.~some statistic of the image labelling. This may take various forms: an image-level label that indicates presence or counts the number of pixels of a particular object category like `sky' or `water', a set of objects' bounding boxes---rectangles that tightly bound object instances' segmentations, or a set of seeds---the pixels that have to take the specified labels~(Fig.~\ref{fig:MSRC-label-types}). More broadly, weakly-supervised learning may be useful in many training problems where the input is obtained by crowdsourcing. For example, some part of a training set for object detection may be of low quality, meaning that the bounding boxes are not tight. In the document tagging problem, low-quality ground truth may miss some tags of the documents. It is preferable to model those biases in the annotation explicitly.


 
As for semantic segmentation, different types of annotations help not only to overcome logistic difficulties, but also to characterize certain categories better. For example, many object categories (i.e.~`things' in terms of \citet{Heitz2008}) are better described by bounding-box annotations, while the background categories (i.e.~`stuff'~\cite{Heitz2008})---which tend to fill significant parts of an image---by image-level labels.


A number of researchers have recognized the importance of weak annotations for learning semantic segmentation. However, most of these methods only use image-level labels. For example,~\citet{Vezhnevets2011,Vezhnevets2012} use a multi-image probabilistic graphical model to propagate image-level annotations across different training images. In this paper, we present a framework for learning structured classification from the mixture of fully and weakly annotated instances. Our framework can employ different types of weak annotations, even for a single instance.

Our work extends recent research on using latent-variable structural support vector machines (\mbox{LV-SSVM}) for weakly-supervised learning~\cite{Chang2010,Kumar2011a,Lou2012} by incorporating \textit{annotation-specific} loss functions, which measure the inconsistency of some labelling predicted by the algorithm with the ground-truth weak annotation. We define those loss functions such that each of them returns an estimate of the expected Hamming loss w.r.t.~all possible labellings consistent with the corresponding weak annotation. Due to this definition, the loss functions specific to different annotation types have the same scale. Our framework thus requires only one coefficient, which balances the relative impact of the loss functions for fully labelled and weakly annotated data, since the latter are typically less informative. We empirically show that balancing between these two kinds of loss functions can improve labelling performance. 

A number of key technical challenges arise while learning an \mbox{LV-SSVM} model with multiple annotation-specific loss functions. These include solution of the \textit{loss-augmented} and \textit{annotation-consistent} inference problems. The former involves finding the labelling that satisfies the current model and deviates from the annotation the most, while the latter involves finding the best labelling that is consistent with the weak annotation. We show how to solve these optimization problems for various loss functions using efficient optimization algorithms. 

\paragraph{Relation to previous work.}
Our work is most closely related to the work of \citet{Kumar2011a}, who use a sequential method to learn semantic segmentation from different types of annotations. Their method starts by training \mbox{LV-SSVM} with a loss function defined on partial labellings; it performs loss-augmented inference using carefully initialized iterated conditional modes (ICM). Once this model is trained, they infer the partial labellings for weakly-annotated images that are consistent with their bounding-box or image-level annotations. The model is then re-trained by considering those solutions as the true partial labellings for the training instances. Unlike \citet{Kumar2011a}, at the training stage we minimize our annotation-specific loss functions simultaneously. In this regard, our framework does not require neither fully nor partially labelled images, which are essential for the first stage of their algorithm. Furthermore, our loss functions allow us to use powerful graph cut based algorithms for solving the loss-augmented  and annotation-consistent inference problems, instead of using an ICM-based inference. Finally, we use different types of weak annotations. 

For some of the loss functions we use, the loss-augmented inference problems cannot be decomposed to the individual variables. This relates us to the recent work on supervised learning with non-decomposable loss functions~\cite{Pletscher2012,Tarlow2012}. \citet{Pletscher2012} use a higher-order loss function that penalizes the difference in the area of the target category between binary segmentations. They show how to use graph cuts for efficient exact loss-augmented inference. \citet{Tarlow2012} use message-passing inference in SSVM training with three different higher-order loss functions: \mbox{PASCAL VOC} loss, bounding box fullness loss, and local border convexity loss.

\paragraph{Our contributions:}~\\[-0.5cm]
\begin{itemize}
\item we propose an \mbox{LV-SSVM} based multi-utility learning framework, which simultaneously minimizes different annotation-specific loss functions, and a unified technique for establishing loss functions for weak annotation of different types;
\item we apply our framework to define the loss functions for training semantic segmentation that are specific to the following weak annotation types and their combinations: image-level labels, bounding boxes, and objects' seeds;
\item we propose efficient inference algorithms required for \mbox{LV-SSVM} training with these loss functions.
\end{itemize}

\section{Latent-variable SSVM}

\subsection{Structured-output learning}
Structured-output learning attempts to learn a mapping $H$ from the space of features~$\mathcal{X}$  to the space of all possible labellings~$\mathcal{Y}$. In what follows, we consider only the mappings that can be expressed as maximization of a discriminant function~$F$ that depends linearly on its parameters~$\vec{w}$:
\begin{equation}
\!\!\! H(\vec{x}) = \argmax_{ \vec{y} \in \mathcal{Y}} F(\vec{x}, \vec{y}; \vec{w}) = \argmax_{ \vec{y} \in \mathcal{Y}} \vec{w}^\intercal  \boldsymbol{\mathrm{\Psi}}(\vec{x}, \vec{y}),
\end{equation}
where vector function~$\boldsymbol{\mathrm{\Psi}}(\vec{x}, \vec{y})$ denotes so-called generalized features of instance~$\vec{x} \in \mathcal{X}$ and labelling~$\vec{y}$. $\boldsymbol{\mathrm{\Psi}}(\vec{x}, \vec{y})$ is defined in a problem-specific way, while the weights~$\vec{w}$ are learned from the training data. We address a wide class of so-called \textit{labelling problems}, where the structured label is a vector of discrete variables: $\mathcal{Y} = \mathcal{K}^V$, where $\mathcal{K} = \{1, \dots, K\}$. Its length $V$ may vary for individual instances.

The goal of supervised structured-output learning is to obtain the most appropriate weights~$\vec{w}$ given the set of features and ground-truth labels of training instances: $\{(\vec{x}_n,\vec{y}_n)\}_{n=1}^N$, $\vec{y}_n \in \mathcal{Y}_n$. Here $\mathcal{Y}_n$ is a set of possible labellings compatible with the $n$-th~instance.
%
In this paper we follow the max-margin formulation of structured-output learning (also called structural support vector machine, SSVM)~\cite{Taskar2004,Tsochantaridis2006,Joachims2009}:
\begin{flalign}\label{eq:ssvm-obj}
&\!\!\! \min_{\vec{w},\boldsymbol{\mathrm{\xi}}\ge\vec{0}} ~\frac{1}{2}\vec{w}^\intercal \vec{w} +\frac{C}{N}\sum_{n=1}^N\xi_n, \\
&\mathrm{s.t.} ~~~ F(\vec{x}_n, \vec{y}_n; \vec{w})\ge 
\max_{\bar{\vec{y}}\in\mathcal{Y}_n} \big(F(\vec{x}_n, \bar{\vec{y}}; \vec{w})+\Delta(\bar{\vec{y}},\vec{y}_n) \big)-\xi_n,\ \ \forall n, \label{eq:ssvm-constr}
\end{flalign}
where $\Delta(\bar{\vec{y}},\vec{y}_n)$ is the loss of some labelling $\bar{\vec{y}}=\{\bar{y_i}\}_{i = 1}^V$ with respect to the ground truth labelling $\vec{y}_n=\{y^n_i\}_{i =1}^{V_n}$. Let $c_i^n$ be some cost associated with the $i$-th~variable in the labelling of the \mbox{$n$-th}~instance. The commonly used loss function is the weighted Hamming distance:
\begin{equation} \label{eq:delta}
\Delta(\bar{\vec{y}},\vec{y}_n)=\sum_{i\in \mathcal{V}_n}c_i^n[\bar{y_i}\ne y^n_i],\footnote{We use the Iverson bracket notation: $[e] = 1$ if the logical expression~$e$ is true, and $[e] = 0$ otherwise}
\end{equation}
This loss function is decomposable w.r.t.~the individual variables. It often implies that loss-augmented inference, i.e.~maximization in~\eqref{eq:ssvm-constr}, is no more difficult than the maximization of discriminant function~$F(\vec{x},\vec{y}; \vec{w})$. In some cases it is possible to use higher-order loss functions that cannot be decomposed w.r.t.~the individual variables~\cite{Pletscher2012,Tarlow2012,Delong2012}.

Problem \eqref{eq:ssvm-obj}--\eqref{eq:ssvm-constr} is convex and can be solved by the cutting-plane method~\cite{Tsochantaridis2006,Joachims2009}. This method replaces the constraint~\eqref{eq:ssvm-constr} with a bunch of linear constraints and then iteratively approximates the feasible polytope by adding the most violated constraint. Such constraint is determined in each iteration by running the loss-augmented inference in~\eqref{eq:ssvm-constr}.

\subsection{Learning with weak annotations} 

Consider the case when in addition to $N$~fully-labelled objects, train set contains $M$~weakly-annotated ones: $\{(\vec{x}_m,\vec{z}_m)\}_{m=N+1}^{N+M}$. Hereinafter we assume that the weak annotation~$\mathbf{z}_m$ defines a subset of full labellings~$\mathcal{L}(\mathbf{z}_m) \subset \mathcal{Y}$ that are consistent with it, and thus~$\mathbf{z}_m$ is less informative than an individual full labelling~$\mathbf{y}_m$. Examples of such weak annotations for the image segmentation problem are (1)~bounding boxes of the segments of a given label; (2)~a value of some global statistic (area, average intensity, number of connected components etc.) for the segments of a given label; (3) subsets of superpixels that belong to a given label (seeds).

We now generalize the standard SSVM formulation to make it handle both fully and weakly annotated data simultaneously. Our multi-utility SSVM is formally defined as follows:
\begin{flalign}
&\!\!\!\!\min_{\substack{\vec{w},  \boldsymbol{\mathrm{\xi}}\ge\vec{0}, \boldsymbol{\mathrm{\eta}}\ge\vec{0}}}
  \frac{1}{2}\vec{w}^\intercal\vec{w} + \frac{C}{N+M}\left(\sum_{n=1}^N\xi_n+\alpha\sum_{m=1}^M\eta_m\right), \label{eq:lvssmv-obj}\\
&\mathrm{s.t.} \quad\quad F(\vec{x}_n, \vec{y}_n; \vec{w})\ge 
 \max_{\bar{\vec{y}}\in\mathcal{Y}_n} \left(F(\vec{x}_n, \bar{\vec{y}}; \vec{w})+\Delta(\bar{\vec{y}},\vec{y}_n) \right)-\xi_n,\ \ \forall n, \label{eq:lvssmv-constr-strong} \\
&\max_{\vec{y} \in \mathcal{L}(\mathbf{z}_m)} F(\vec{x}_m, \vec{y}; \vec{w})\ge 
 \max_{\bar{\vec{y}}\in\mathcal{Y}_m} \left(F(\vec{x}_m, \bar{\vec{y}}; \vec{w})+ \Kappa(\bar{\vec{y}},\vec{z}_m) \right)-\eta_m, \forall m. \label{eq:lvssmv-constr-weak}
\end{flalign}

Note that for $M=0$ the above formulation degenerates to the standard SSVM formulation, while for $N=0$ it reduces to the latent-variable SSVM~\cite{Yu2009}. Note also that the full labelling $\vec{y}_n$ can be seen as a degenerate weak annotation, where $\mathcal{L}(\mathbf{z}_m) = \{\vec{y}_n$\}. Therefore, Problem~\eqref{eq:lvssmv-obj}--\eqref{eq:lvssmv-constr-weak} is almost equivalent to \mbox{LV-SSVM}, but it contains the slack balancing coefficient $\alpha$. Ignoring this coefficient may hurt the performance of multi-utility learning, as we show in Section~\ref{sec:res-imagelevel}. In order to perform the optimization, in addition to the loss-augmented inference in~\eqref{eq:lvssmv-constr-strong}, we should also be able to perform the weak-loss augmented inference in~\eqref{eq:lvssmv-constr-weak}, as well as the \textit{annotation-consistent inference} in the left-hand side of~\eqref{eq:lvssmv-constr-weak}.


Optimization problem~\eqref{eq:lvssmv-obj}--\eqref{eq:lvssmv-constr-weak} is not convex and thus hard. We follow~\citet{Yu2009} and use the concave-convex procedure (CCCP)~\cite{Yuille2002} to solve it approximately.

\section{Weak annotation for semantic image segmentation} \label{sec:weaklabel}

Semantic image segmentation aims to assign category labels to image pixels. We assume that an image is represented as a set of \textit{superpixels}, i.e.~groups of co-located pixels similar by appearance. Consider a graph~$\mathcal{G}=(\mathcal{V},\mathcal{E})$. Its nodes $\mathcal{V}$ correspond to superpixels of the image. The set of edges~$\mathcal{E}$ represents a neighborhood system on $\mathcal{V}$ that includes the pairs of nodes that correspond to all adjacent superpixels. Let $\vec{x}_i\in\mathbb{R}^d$ be a vector of superpixel features associated with some node~$i \in {\mathcal V}$, $\vec{x}_{ij}\in\mathbb{R}^e$ be a vector of superpixel interaction features for the edge connecting nodes~$i$ and~$j$, and $\vec{x}= \bigoplus_{i\in\mathcal{V}} \vec{x}_i \oplus \bigoplus_{(i,j)\in\mathcal{E}}\vec{x}_{ij}$ be their concatenation. The value of each variable~$y_i$ corresponds to the label of the $i$-th superpixel. We use the following discriminant function~$F$:
\begin{equation} \label{eq:energy}
F(\vec{x}, \vec{y}; \vec{w}) = \vec{w}^\intercal  \boldsymbol{\mathrm{\Psi}}(\vec{x}, \vec{y}) =
\sum_{i\in \mathcal{V}}\sum_{k=1}^K [y_{i} = k] (\vec{x}_i^\intercal \vec{w}_k^\mathrm{u}) +
\!\!\! \sum_{(i,j)\in \mathcal{E}} \![y_{i} = y_{j}] (\vec{x}_{ij}^\intercal \vec{w}^\mathrm{p}),
\end{equation}
where $\vec{w}=\bigoplus_{k=1}^K \vec{w}_k^\mathrm{u} ~\oplus \vec{w}^\mathrm{p}$ is a vector of the model parameters, and~$\vec{w}_k^\mathrm{u}\in\mathbb{R}^d$, $\vec{w}^\mathrm{p}\in\mathbb{R}^e$ . The summands in the first and the second terms are called unary and pairwise potentials, respectively. We restrict pairwise weights~$\vec{w}^\mathrm{p}$ and pairwise features $\vec{x}_{ij}$ to be nonnegative and thus obtain an associative discriminative function (with only attractive pairwise potentials)~\cite{Taskar2004}. Maximizing $F(\vec{x},\vec{y}; \vec{w})$ w.r.t.~$\vec{y}$ is known to be NP-hard, but efficient approximate algorithms exist, e.g.~$\alpha$-expansion~\cite{Boykov2001}.

We use the weighted Hamming loss~\eqref{eq:delta} for fully-labelled images, where $c_i$ is the number of pixels in the corresponding superpixel, so the loss function estimates the number of mislabelled image pixels.\footnote{In practice, ground-truth labelling of a superpixel may contain several labels; in this case the number of incorrectly inferred pixels is added to the loss. We ignore this case to ease the notation, but all the algorithms still work in that case.} To use some type of weak annotations for training, we need to define the annotation-specific loss function that allows loss-augmented inference and annotation-consistent inference. The former should be efficient, since it is performed in the inner loop of training and thus is typically a bottleneck. We show how to define and combine them for the annotations of the following types: image-level labels, bounding boxes around objects, and objects' seeds.

\subsection{Image-level labels}
We start by defining loss functions~$\Kappa(\vec{y},\vec{z})$ for some arbitrary labelling $\vec{y}$ and ground-truth weak annotation $\vec{z}$. In this subsection we assume that $\vec{z}$ is a set of labels used in the ground-truth image labelling (for the image in Fig.~\ref{fig:MSRC-label-types}, $\vec{z} = \{$`sky', `tree', `plain', `grass'\}). We cannot compute the Hamming loss \eqref{eq:delta} if the full labelling is unknown for one of its arguments. Let's instead define a proxy loss function, that is symmetric and does not compare labels of any superpixels directly:
\begin{equation} \label{eq:delta-ill}
\Delta_{\textrm{il}}(\vec{y}, \bar{\vec{y}}) = 
\sum_{i \in \mathcal{V}}c_i[\nexists j \in \mathcal{V}: y_j = \bar{y}_i ~\lor~ \nexists j \in \mathcal{V}: \bar{y}_j = y_i].
\end{equation}
It penalizes all the superpixels that have been given any label that lacks in the annotation~$\bar{\vec{y}}$, as well as superpixels which have ground truth labels that lack in~$\vec{y}$. Unfortunately, the ground-truth labelling~$\bar{\vec{y}}$ is unknown. If we knew the areas $S_k$ of each label $k \in \vec{z}$, we could derive the following upper bound on~\eqref{eq:delta-ill}:
\begin{equation} \label{eq:upp-bnd-kappa}
\Kappa_{\textrm{il}}(\vec{y}, \vec{z}; \{S_k\}_{k \in \mathbf{z}}) = 
\sum_{k \not\in \vec{z}} \sum_{i \in \mathcal{V}}c_i[y_i = k] +
\sum_{k \in \vec{z}} S_k \prod_{i\in \mathcal{V}}[y_i\ne k].
\end{equation}
This upper bound is tight up to a factor of 2. The first term penalizes the pixels labelled with wrong labels, while the second term penalizes ignoring the labels from $\vec{z}$.

Since we do not know the areas $S_k$, the best we can do is to assume~$\Kappa(\vec{y},\vec{z})$ to be the expectation of \eqref{eq:upp-bnd-kappa} taken over all full labellings consistent with $\vec{z}$. If there are enough fully-labelled images, the areas $S_k$ can be estimated. Otherwise we assume the uniform distribution over the feasible full labellings $\vec{y} \in \vec{z}$ and get
\begin{equation} \label{eq:kappa-il}
\Kappa_{\textrm{il}}(\vec{y}, \vec{z}) =
\sum_{k \not\in \vec{z}} \sum_{i\in \mathcal{V}}c_i[y_i = k] +
\sum_{k \in \vec{z}} \frac{\sum_{i \in \mathcal{V}} c_i}{|\vec{z}|} \prod_{i\in \mathcal{V}}[y_i\ne k].
\end{equation}

Having defined the loss function $\Kappa_{\mathrm{il}}$, we need to provide algorithms for inference problems in~\eqref{eq:lvssmv-constr-weak}. 
For annotation-consistent inference $\max_{\vec{y} \in \vec{z}_m} F(\vec{x}_m, \vec{y}; \vec{w})$ we use $\alpha$-expansion over the labels from $\vec{z}_m$ only. Note that this may result in an inconsistent labelling: some labels from $\vec{z}_m$ may miss in $\vec{y}$. We have tried an heuristic algorithm for making it strictly consistent with $\vec{z}$ by changing one node per missing label, but observed no significant difference in practice.

The loss-augmented inference is now not decomposable to unary and pairwise factors. To work this around, we derive:
\begin{multline} \label{eq:lai-il}
\!\!\!\max_{\bar{\vec{y}}\in\mathcal{Y}_m} \left(F(\vec{x}_m, \bar{\vec{y}}; \vec{w})+K_{\textrm{il}}(\bar{\vec{y}},\vec{z}_m) \right) = \\
\max_{\bar{\vec{y}}\in\mathcal{Y}_m} \Bigg(F(\vec{x}_m, \bar{\vec{y}}; \vec{w})+
  \sum_{k \not\in \vec{z}} \sum_{i\in \mathcal{V}}c_i[\bar{y_i} = k] - \sum_{k \in \vec{z}} \frac{\sum_{i \in \mathcal{V}} c_i}{|\vec{z}|} [\exists i : \bar{y_i} = k] \Bigg) + \textrm{const}.
\end{multline}
The last maximization is the standard MRF inference problem with label costs. We use the efficient modification of $\alpha$-expansion for accounting label costs~\cite{Delong2012}.

\subsection{Bounding boxes}
It is convenient to annotate instances in an image with tight bounding boxes (Fig.~\ref{fig:MSRC-bbox}). On the other hand, they do not give much information for background categories. Therefore, we consider the annotation that consists of both bounding boxes and image-level labels. For example, annotation of an image may contain the bounding-box locations of cars and pedestrians, and additionally state that there are buildings, road, and sky in the image. We assume that within a certain image each category can be defined either with image-level labels, or with bounding boxes, though the type of annotation for a category may vary from image to image (see Section~\ref{sec:bbox} for an example where it can be useful).

We model weak annotation $\vec{z}$ of an image as a pair $(\vec{z}^{\textrm{il}}, \vec{z}^{\textrm{bb}})$ of image-level and bounding box annotations. The latter is a set of bounding boxes with associated category labels: $ \vec{z}^{\textrm{bb}} = \{z_i\}$, with the following functions defined: $\textit{label}(z_i)$, which defines the associated category label, and $\textit{box}(z_i) = [\textit{left}(z_i), \textit{right}(z_i)] \times [\textit{top}(z_i), \textit{bottom}(z_i)]$ that defines the extent of the bounding box. The set of labels $\mathcal{K}$ is partitioned into three subsets w.r.t.~the weak annotation~$\vec{z}$: the labels that are defined with bounding boxes~($\mathcal{K}_b = \bigcup_{z \in \vec{z}^{\textrm{bb}}} \textit{label}(z)$), those that are present somewhere else in the image~($\mathcal{K}_p = \vec{z}^{\textrm{il}}$), and those that are absent~($\mathcal{K}_a = \mathcal{K} \setminus (\mathcal{K}_b \cup \mathcal{K}_p)$). Nodes $\mathcal{V}$ are also partitioned: $\mathcal{V}_k =  \bigcup_{\substack{z \in  \vec{z}^{\textrm{bb}}:  \textit{label}(z) = k}} \textit{box(z)}$ is the union of pixel indices in the bounding boxes corresponding to the label $k \in \mathcal{K}_b$, and $\mathcal{V}_0 = \mathcal{V} \setminus \bigcup_{k \in \mathcal{K}_b} \mathcal{V}_k$. We now define the combined loss function as: 
\begin{multline} \label{eq:kappa-il-bb}
\!\!\!\!\!\!\!\Kappa_{\textrm{il-bb}}(\vec{y}, \vec{z}) =
\sum_{k \in \mathcal{K}_a} \sum_{i \in \mathcal{V}} c_i \left[y_i = k\right] +
\sum_{k \in \mathcal{K}_p} \sigma_k \prod_{i \in \mathcal{V}} \left[y_i \ne k\right] + \\
\beta \sum_{z \in \vec{z}^{\textrm{bb}}} \Bigg( \sum_{p = \textit{top}(z)}^{\textit{bottom}(z)} \!\!\nu^z_{p}\!\!\!\! \prod_{q = \textit{left}(z)}^{\textit{right}(z)}\!\!\!\!\!\! \textit{V}((p,q);\vec{y},\textit{label}(z)) \,\, + 
\sum_{q = \textit{left}(z)}^{\textit{right}(z)} \!\!\omega^z_{q}\!\!\!\! \prod_{p = \textit{top}(z)}^{\textit{bottom}(z)}\!\!\!\!\!\! \textit{V}((p,q);\vec{y},\textit{label}(z)) \Bigg) \\
+ \sum_{k \in \mathcal{K}_b} \sum_{i \in \mathcal{V}_0} c_i \left[y_i = k\right].
\end{multline}
The first two terms are almost the same as in \eqref{eq:kappa-il}, but the estimate of the category area in the second term does not include the pixels within the bounding boxes: $\sigma_k = \left(\sum_{i \in \mathcal{V}_0} c_i \right) / |\vec{z}^\textrm{il}|$. The third term penalizes `empty' rows and columns in the bounding boxes, i.e.~those rows and columns that do not contain pixels of a target category at all. The violation function $V$ is defined as:
\begin{equation}
\textit{V}(\vec{p};\vec{y},k) = \begin{dcases*}
1, & if $\textit{map}(\vec{y})_{\vec{p}}\ne k$,\\
0, & otherwise.
\end{dcases*}
\end{equation}
Here $\textit{map}(\vec{y})$ is the classification map induced by the superpixel labelling $\vec{y}$. Coefficients~$\nu^z_p$ and~$\omega^z_q$ allow us to assign the penalty for the corresponding row or column being empty, depending on its position in the bounding box. One can learn the category-specific profiles of $\nu^z$ and $\omega^z$ when the full labelling is abundant enough, but we use uniform profiles assuming that half of a bounding box is occupied by the object on average: $\nu^z_p = \big(\textit{right}(z) - \textit{left}(z) \big) / 2$, $\omega^z_q = \big(\textit{bottom}(z) - \textit{top}(z) \big) / 2$. Note that this makes the loss an estimate on the number of mislabelled pixels (similar to the image-level label loss~\eqref{eq:kappa-il}), so the value coefficient $\beta = 1$ should work well (we show in Section~\ref{sec:bbox} that it really does). We have also tried linearly decreasing loss used by \citet{Kumar2011a}, but it did not affect the performance significantly. The last term penalizes the bounding-box labels outside of bounding boxes.

We have shown in the previous section how to account for the two initial terms in the loss-augmented inference. The last term is decomposable w.r.t.~superpixels. The third term is a sum over the higher-order cliques of the following form. For each bounding box $z$, each row and each column generates a clique of nodes corresponding to the superpixels that intersect that row/column. We treat them the same way as the image-level loss: we modify $\alpha$-expansion with label costs~\cite{Delong2012} to penalize each clique of superpixels, which contains at least one superpixel labelled with $\textit{label}(z)$. There is a technical difficulty with the superpixels that cross the bounding box border: it is unclear if their labelling with $\textit{label}(z)$ should be penalized. We adopted the following strategy: shrink the bounding box to allow some margin, and treat all superpixels that intersect the shrunk bounding box (and only them) as insiders. We set the margin width equal to 6\% of the corresponding bounding box dimension.

During the annotation-consistent inference, we need to infer a labelling that has objects only in bounding boxes of the corresponding category labels, and they should fill those bounding boxes tightly, i.e.~touch upon all four sides of the bounding box shrunk to allow a 6\% margin (\citet{Lempitsky2009} showed that this definition is natural). The first condition is easy to satisfy: we can suppress certain labels outside of bounding boxes by using infinite unary potentials. To provide tightness, we use a variation of the pinpointing algorithm~\cite{Lempitsky2009}, adapted for the multi-class segmentation. First, segmentation is performed without the tightness constraints. Then, until those constraints are satisfied, one of the superpixels changes its unary potential, and expansion move is performed. In our implementation, we select the superpixel with the highest relative potential for $\textit{label}(z)$ that has not been assigned this label yet, and assign it the infinite potential for $\textit{label}(z)$ to guarantee that it will change its label. This procedure is finite because at each iteration at least one superpixel within $\textit{box}(z)$ switches to $\textit{label}(z)$. In contrast to~\citet{Lempitsky2009}, we do not perform further dilation, since it is unclear, which label we should use for expansion move(s); neither of the heuristics we tried improved the result significantly. We also found that initialization of the latent variables in \mbox{LV-SSVM} matters: we obtained the best results when initially all superpixels within $\textit{box}(z)$ were initialized with $\textit{label}(z)$. Note that~\citet{Kumar2011a} used a different criterion during the annotation-consistent inference: they penalize the empty rows and columns within bounding boxes (the opposite to what we do in loss-augmented inference). Note that their heuristic does not guarantee the tightness of the resulting segmentation.

\subsection{Objects' seeds}
Another form of a weak annotation natural for the object categories is the seed annotation (Fig.~\ref{fig:MSRC-seed}). In general, for a segment of some category, a seed is a subset of its pixels. We consider a particular case, where only one pixel, persumably close to the segment center, is labelled. During the annotation-consistent inference, we require the superpixel where this point is located to have the fixed seed label.

We now model the weak annotation $\vec{z}$ as a pair $(\vec{z}^{\textrm{il}}, \vec{z}^{\textrm{os}})$, where $\vec{z}^{\textrm{os}}$ is a set of 2D points with the corresponding labels: $(\vec{p}, k)$. The seed centrality assumption allows us to set the Gaussian penalty for inferring any non-seed label in the neighbourhood of each seed, which brings us to the following loss function:
\begin{multline} \label{eq:kappa-il-os}
\Kappa_{\textrm{il-os}}(\vec{y}, \vec{z}) =
\sum_{k \in \vec{k}_a} \sum_{i \in \mathcal{V}} c_i \left[y_i = k\right] +
\sum_{k \in \vec{k}_p} \sigma_k \prod_{i \in \mathcal{V}} \left[y_i \ne k\right] + \\[-0.25cm]
\beta \sum_{\substack{(\vec{p}', k') \\ \in \vec{z}^{\textrm{os}}}} \sum_{\vec{p} \in I} \mathit{V}(\vec{p}; \vec{y}, k') \exp \left(-\frac{\pi \| \vec{p} - \vec{p}' \|^2} {\tau_{k'}} \right).
\end{multline}
Here the first two terms are the same as in the image-level label loss. The inner sum in the third term is taken over all image pixels $I$. The form of the Gaussian is defined in such a way that the penalty for misclassification of the central pixel $\vec{p'}$ is 1, and whenever no superpixels of the label~$k'$ are found, the penalty is equal to the estimated area of the label~$k'$ w.r.t.~all labellings consistent with the weak annotation; specifically, 
\begin{equation}
\!\!\tau_{k'} = \frac{\sum_{i \in \mathcal{V}} c_i}
{\left(|\vec{z}^{\textrm{il}}| + \textit{\#Lab}(\vec{z}^{\textrm{os}}) \right) \cdot \textit{\#Obj}(\vec{z}^{\textrm{os}}, k')}.
\end{equation}
Here $\textit{\#Lab}(\vec{z}^{\textrm{os}})$ is the number of different labels in $\vec{z}^{\textrm{os}}$, and $\textit{\#Obj}(\vec{z}^{\textrm{os}}, k')$ is the number of seeds of the label $k'$ in $\vec{z}^{\textrm{os}}$. Loss~\eqref{eq:kappa-il-os} is decomposable to factors, so the loss-augmented inference is trivial.

\section{Experiments}

\subsection{Datasets and metrics}
We test the proposed framework on two datasets: MSRCv2\footnote{\url{http://research.microsoft.com/en-us/projects/objectclassrecognition/}} \cite{Shotton2006,Vezhnevets2011} and SIFT-flow\footnote{\url{http://people.csail.mit.edu/celiu/LabelTransfer/code.html}} \cite{Liu2009,Tighe2010,Vezhnevets2012}. MSRC contains 276 training and 256 test images that are fully labelled using 23 category labels; significant part of pixels remains unlabelled. SIFT-flow is a more challenging dataset: it is a subset of the LabelMe database~\cite{Torralba2010}, which contains 2488 training and 200 test images; they are labelled to 33 categories using crowd-sourcing. 

For \textit{MSRC}, we obtain superpixels using the original implementation of the \textit{gPb} edge detector~\cite{Arbelaez2011}. The unary features are the following: a histogram of SIFT~\cite{Lowe2004} visual words built using a dictionary of size 512 by hard assignment of the descriptors to the bins; a histogram of the RGB colors on a dictionary of size 128; a histogram of locations over a uniform $6 \times 6$ grid. We $L_2$-normalize the joint feature vector and map it into a higher-dimensional space where the inner product approximates the $\chi^2$-kernel in the original space (the dimensionality of the space triples after the transformation)~\cite{Vedaldi2010}. We use pairwise factors over the pairs of the superpixels that share a common border and use the following 4 pairwise features: $\exp (-c_{ij}/10)$, $\exp (-c_{ij}/40)$, $\exp (-c_{ij}/100)$, 1. Here $c_{ij}$ is the strength of the boundary between segments $i$ and $j$ returned by the \textit{gPb}.

For \textit{SIFT-flow}, we follow \citet{Vezhnevets2012} and obtain superpixels and features using the code by \citet{Tighe2010}. It runs graph-based segmentation of~\citet{Felzenszwalb2004} followed by feature extraction. The unary features include shape, location, texture, color and appearance feature vectors, some of which are also computed over dilated superpixel masks to capture the context: 3115 unary features in total. We also transform this vector with a~$\chi^2$-kernel approximation, which triples its size. We use pairwise factors over the pairs of superpixels that share a common border and the pairwise features computed as distances between groups of superpixels' features ($\chi^2$ distance in case of histograms, Euclidean otherwise), 26 features in total.


\paragraph{Quality measures.} 
We use two standard measures of segmentation quality: accuracy and per-class recall. The accuracy is defined as the rate of correctly labelled pixels of the test set. The per-class recall is the number of correctly labelled pixels of each category divided by the true total area of that category, averaged over categories. Following the previous work \cite{Vezhnevets2011,Shotton2008}, we exclude the pixels of rare categories (`horse' and `mountain') from recall computation for MSRC, but include the `other' label, see Section~\ref{sec:res-imagelevel}. Similarly, we exclude rare categories (`cow', `desert', `moon', and `sun') from SIFT-flow recall computation.


\subsection{Image-level labels}\label{sec:res-imagelevel}

\paragraph{Generating weak annotation.} 
We obtain image-level labels automatically from full labellings by enumerating the unique labels for each image. Each MSRC image typically features one or several objects of some target category (e.g.~`sign', `cow', `car') on top of some background. Not every background category falls into the used labels, so it may remain unlabelled. Thus, some images contain only one category label. In this case the image-level label unambiguously defines the full labelling. To avoid this knowledge (unrealistic in the real-world setting), we could model the `other' label, which contains anything but the labelled 23 categories. However, the labellings typically have uncertain borders between segments of different labels, i.e.~the borders are unlabelled too (Fig.~\ref{fig:MSRC-strong}). If we modelled those boundaries as a separate category, it would hurt the segmentation performance. Instead, we want to model this `other' label only for unlabelled regions, not for the boundaries. We use the following heuristic criterion for obtaining image-level labels: if an image contains only one label, or at least 30\% of its pixels are unlabelled, we include them to the image-level label as the `other' label.

\begin{figure*}[tb]
\begin{center}
    \begin{minipage}[b]{.328\linewidth}
        \includegraphics[scale=0.666]{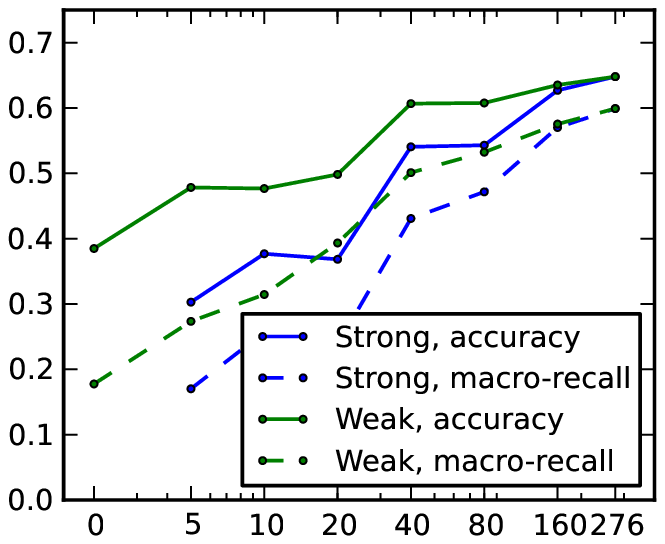}
        \\[-0.7cm]\subcaption{}
        \label{fig:MSRC-B01}
    \end{minipage}
    \begin{minipage}[b]{.328\linewidth}
        \includegraphics[scale=0.666]{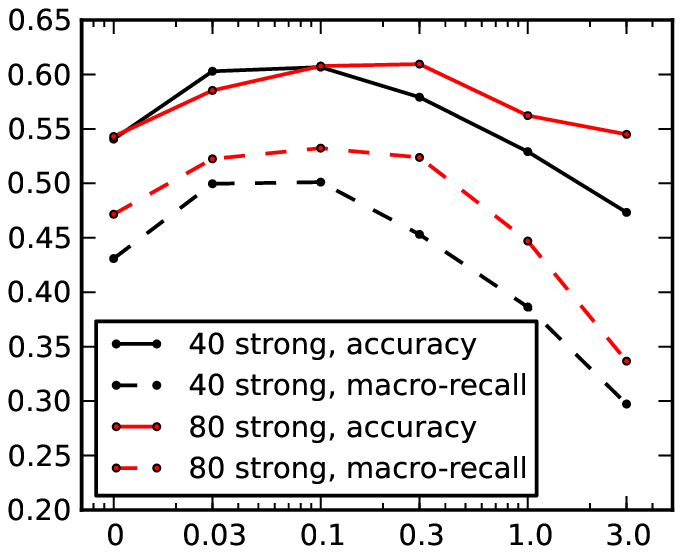}
        \\[-0.7cm]\subcaption{}
        \label{fig:MSRC-varcoef}
    \end{minipage}
    \begin{minipage}[b]{.328\linewidth}
        \includegraphics[scale=0.666]{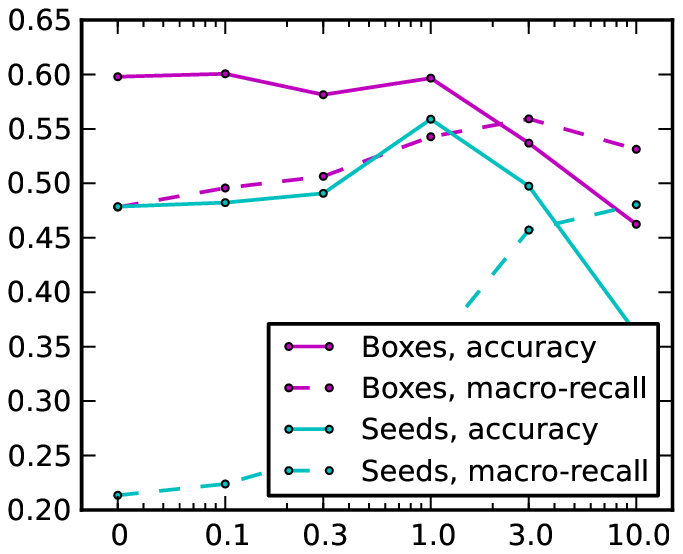}
        \\[-0.7cm]\subcaption{}
        \label{fig:MSRC-varcoefBboxSeed}
    \end{minipage}
    \\[-0.2cm]\caption{ (a)--(c) Accuracy (solid lines) and per-class recall (dashed lines) subject to different parameters on the MSRC dataset. (a) Varying the number of fully-labelled images. Blue line show test set segmentation quality when only fully-labelled images are available; green line---when the complementary part of the train set has image-level labels. (b) Varying the coefficient of the weak-loss coefficient $\alpha$. Black line show test set segmentation quality when 40 images are fully labelled, red line---when 80 images; the complementary part of the train set has image-level labels. (c) Varying the coefficient of the bounding box (magenta line) or object seed (cyan line) loss $\beta$. All 276 training images have image-level labels, all objects have tight bounding box or seed annotations, respectively\\[-5ex] }
    \label{fig:MSRC-res}
\end{center}
\end{figure*}

\begin{table}[t]
    \begin{center}
\makebox[0pt][c]{\parbox{1.0\linewidth}{
   \begin{minipage}[b]{0.42\linewidth}\centering
	\caption{Accuracy and average per-class recall on the SIFT-flow dataset. The first two lines describe training on the subset of 256 fully labelled images of the models with and without pairwise potentials, respectively. The third line experiment used the whole dataset with image-level labels, but for only 256 of them full labelling is known. The bottom line shows the result when the whole dataset is fully labelled\\[-0.05cm] }\label{tab:SIFT-flow-res} \small
	\begin{tabular}{|l|c|c|}\hline
experiment & acc & rec\\ \hline
256/256 strong, local & 0.574 & 0.167\\
256/256 strong, init loc. & 0.620 & 0.176\\
256/2488 strong, init $\uparrow$& 0.674 & 0.208\\
2488/2488 strong & 0.696 & 0.246\\ \hline
	\end{tabular}
   \end{minipage}
   \hfill \centering 
   \begin{minipage}[b]{0.56\linewidth}
	\caption{Accuracy (first number in each cell) and average per-class recall (second number) on the MSRC dataset when during training i) only full labelling is available, ii) image-level (il) labels are also available for the rest of the data set, iii) object seeds (os) are additionally available, iv) bounding boxes (bb) for objects are available, v) both seeds and bounding boxes are available. Note that the numbers in the last column are all equal since the weak annotation does not add any information when all training set is fully labelled\\[-0.4cm] }\label{tab:MSRC-bbox-res} \small
	\begin{tabular}{|ccc|c|c|c|}\hline
il & bb & os & 0/276 strong & 5/276 strong & 276 strong\\ \hline
$-$ & $-$ & $-$ & n/a                     & 0.300/0.170 & 0.648/0.599\\
$+$ & $-$ & $-$ & 0.385/0.178 & 0.478/0.273 & 0.648/0.599\\
$+$ & $-$ & $+$ & 0.559/0.346 & 0.574/0.370  & 0.648/0.599\\
$+$ & $+$ & $-$ & 0.597/0.543 & 0.606/0.546 & 0.648/0.599\\
$+$ & $+$ & $+$ & 0.531/0.567 & 0.542/0.564 & 0.648/0.599\\\hline
	\end{tabular}
	\\[-6.1ex]
  \end{minipage}
}}
   \end{center}
\end{table}

\paragraph{Varying the full-labelling rate.} 
In our basic setting we have a (possibly empty) part of the training set fully labelled, while the rest of the images have only image-level labels. We generate those subsets using the Metropolis--Hastings sampling, trying to make the distribution of their label counts approximate that of the whole training set. 
Fig.~\ref{fig:MSRC-B01} shows the accuracy and per-class recall of the test set segmentation for various full labelling rates in comparison to the fully-supervised setting.\footnote{\url{http://shapovalov.ro/data/MSRC-weak-train-masks.zip}
} In the most common scenario---when less than 20\% of the training set is fully labelled---the weakly-annotated subset provides a stable 10--15\% improvement both in terms of the accuracy and mean per-class recall. 


\paragraph{Balancing the loss functions.} 
When the training set consists of both weak annotations and full labellings, the coefficient~$\alpha$ from~\eqref{eq:lvssmv-obj} needs to be set. We discovered that its optimal value was lower than 1 (Fig.~\ref{fig:MSRC-varcoef} shows the dependency of performance on~$\alpha$). We speculate that this is because we are more certain about the strong loss, so it should contribute to the objective more. Thus, for all the other experiments we set $\alpha=0.1$.

\paragraph{SIFT-flow results.} 
On the SIFT-flow dataset, we compare fully-supervised learning with weakly-supervised at one point, i.e.~when only 256 training images are fully labelled, and the rest 2232 images have only image-level labels (Table~\ref{tab:SIFT-flow-res}). This weakly-learned model loses to the fully-supervised one only 2\% in the accuracy and 4\% in the per-class recall. Note that our model is \textit{on par} with \citet{Vezhnevets2012}, who also reached 21\% on that dataset with the same superpixels and features. The difference is they used only image-level annotation, while we used about 10\% fully labelled images. However, their model is substantially more complicated: they use extremely-randomized hashing forest for non-linear feature transform, learn objectness and image-level priors, and connect superpixels of different images within the multi-image model.
Since the \mbox{LV-SSVM} optimization problem is not convex, the algorithm may get stuck at local minima. We initialize the parameters of \mbox{LV-SSVM} by the parameters of the SSVM trained on the fully-labelled part of the dataset, if there is one. 


\subsection{Adding bounding boxes and seeds} \label{sec:bbox}
\paragraph{Generating weak annotation.} 
We generate two more annotations for the MSRC training data to test additional annotation-specific loss functions. Similar to image-level labels, we generate them from the full  labelling. Tight bounding boxes and object seeds are good for description of the object (`thing') categories, while do not add much information beyond image-level labels for the background (`stuff') categories. We divide the list of categories into two parts: background, which includes `grass', `sky', `mountain', `water', `road', and `other'; and objects, which includes all other categories. There are two ambivalent categories---`building' and `tree'---which can instantiate either a target object of a photograph, or background. We used the following heuristic for each image: consider tree and building as background iff there are other objects in the image. We enhanced the image-level labelling with either tight bounding boxes or object seeds for segments of object categories only. For the other categories, only image-level labels were available. To generate seeds, for each segment we took its pole of inaccessibility---the point that maximizes its distance transform map.
%

\paragraph{Results.} 
Table~\ref{tab:MSRC-bbox-res} summarizes the results. When the full labelling is unavailable, both object seed and bounding box annotations give significant improvement over just image-level labels. Bounding boxes notably increase per-class recall: they help to better learn `thing' categories, which are numerous and typically have smaller area. Overall, learning with bounding boxes only 5\% inferior to learning on fully labelled data both in terms of the accuracy and per-class recall. Object seed annotation gave more modest increase in performance, though is easier to obtain. We used the value $\beta = 1$ to balance the impact of image-level vs.~bounding box (or seed) loss functions: they seem to provide equal contribution to the objective function; Fig.~\ref{fig:MSRC-varcoefBboxSeed} supports that hypothesis.

\begin{figure}[t!]
\begin{center}
        \includegraphics[width=.9\linewidth]{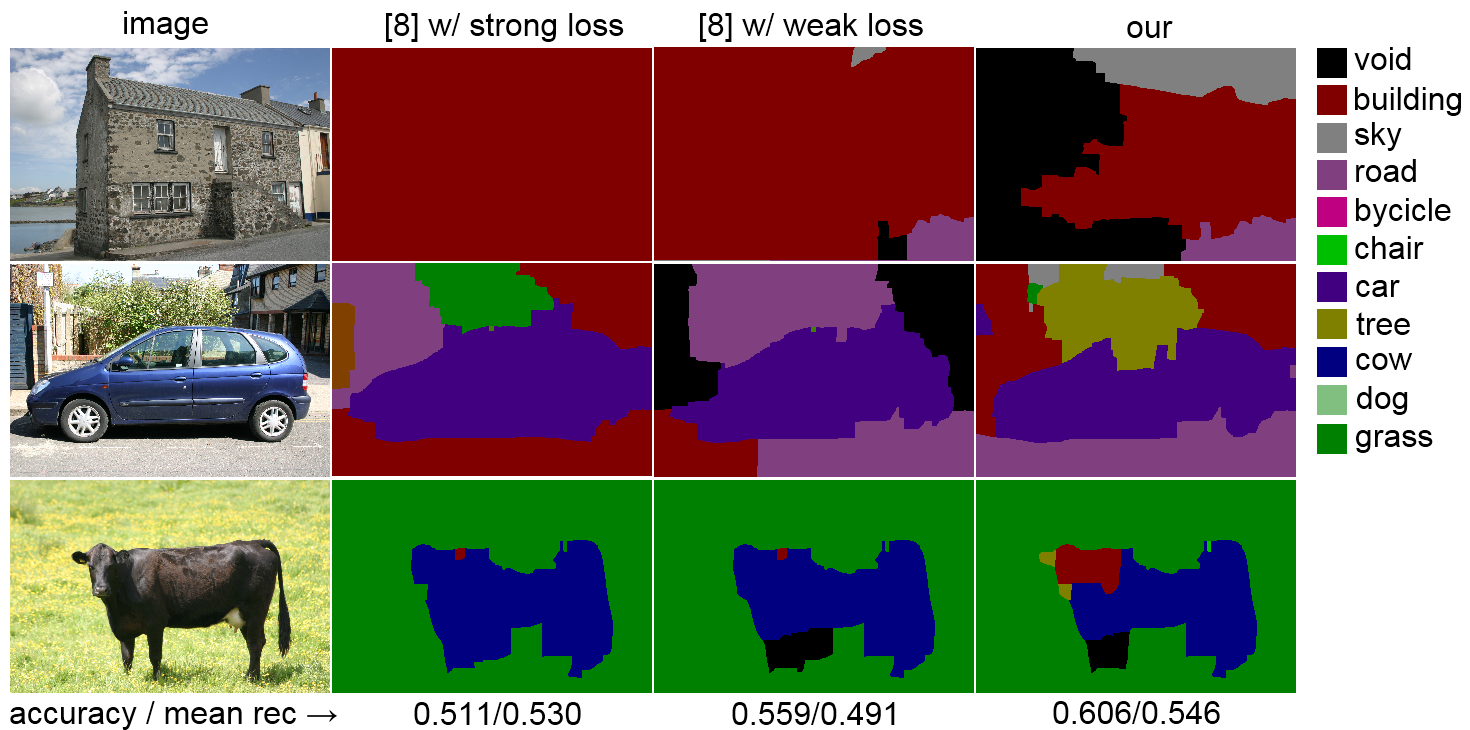}
    \caption{Qualitative results of the proposed algorithm and two variations of the algorithm by \citet{Kumar2011a} applied to three images from the MSRC test set\\[-5ex]}
    \label{fig:MSRC-qualit}
\end{center}
\end{figure}
\paragraph{Comparison to \citet{Kumar2011a}.}
Unfortunately, we cannot directly compare to \citet{Kumar2011a} since the type of input data for their framework is unorthodox. They use two different datasets to obtain segmentation maps (partial labellings) for the foreground and background categories, respectively. Our framework does not support this kind of annotation: we believe that it is easier to obtain segmentation for background and foreground categories using the same set of images. This eliminates the need to use the latent-variable SSVM for training the basic model; instead the global minimum of SSVM objective can be found efficiently. Also, when both image-level labels and bounding boxes (or seeds) are known for each weakly-annotated image, both background and foreground partial labellings can be inferred, and using latent-variable SSVM after adding weakly-annotated data is not necessary again. Thus, when given the data we use, the method of \citet{Kumar2011a} could look like this:\\[-4ex]
\begin{itemize}
\item train SSVM using the fully-labelled part of the training set,
\item use the trained model to infer the labelling of all images consistent with the weak annotation,
\item train SSVM using the hallucinated labelling obtained in the previous step.
\end{itemize}
This method is similar to running one outer iteration of our training algorithm, but it has one important difference: the loss function in the second SSVM. While our method uses the weak loss function, the modified method of \citet{Kumar2011a} uses the strong loss function w.r.t.~the hallucinated labelling. To compare the methods, we use the MSRC training set with 5 fully-labelled images and the rest annotated with bounding boxes and image-level labels~(row 4, column 2 in Table~\ref{tab:MSRC-bbox-res}, excluding headers) to train both described modifications: with the weak bounding-box loss function~\eqref{eq:kappa-il-bb}, and with the strong loss function~\eqref{eq:delta}~(still different from the loss function of \citet{Kumar2011a}). The segmentation maps and numerical results in Fig.~\ref{fig:MSRC-qualit} show that the proposed simultaneous minimization of loss functions is superior both in terms of accuracy and per-class recall. 

\section{Conclusion}
We presented the framework for learning structural classification from different types of annotations by minimizing annotation-specific loss functions. We applied it to semantic image segmentation by introducing weak loss functions for for image-level, bounding box, and object seed annotations. Usage of weakly-annotated training data consistently improves the labelling. The results on the semantic segmentation datasets show that the joint annotation where background is given by image-level labels, and objects are given by bounding boxes, is the best trade-off between segmentation quality and annotation effort.

%


\begingroup
\renewcommand{\section}[2]{\subsubsection*{References}}%
{\small
\bibliographystyle{apalike}
\bibliography{multi-utility}
}
\endgroup

\end{document}